\documentclass[letterpaper, 10 pt, conference]{ieeeconf}
\IEEEoverridecommandlockouts
\usepackage{graphicx} % img insert
\usepackage{subcaption} % subfigure
\usepackage[dvipsnames*,svgnames*]{xcolor}
\usepackage{transparent}
\usepackage{amsmath} % align equation

\usepackage{colortbl}
\usepackage{booktabs}
\usepackage{tabularx}
\usepackage{tabularray}
\usepackage{adjustbox}
\usepackage{multirow}
\usepackage{siunitx}
\usepackage{amsfonts}
\usepackage{xcolor}

\usepackage{siunitx}
\usepackage{booktabs}
\usepackage{xcolor}
\usepackage{array}

\usepackage{textcomp,gensymb}
\usepackage[normalem]{ulem} 
\definecolor{RoyalBlue}{rgb}{0.25,0.41,0.88}

\makeatletter
\let\NAT@parse\undefined
\makeatother
\usepackage[hidelinks]{hyperref} %hide the borders of the cross-references
\usepackage{url}

\usepackage{tikz}
\usetikzlibrary{shadows}
\usetikzlibrary{shadows.blur}
\usetikzlibrary{arrows.meta}
\usepackage[skins]{tcolorbox}

\begin{document}
\title{\LARGE \bf Ensemble of Pre-Trained Models for Long-Tailed Trajectory Prediction}
\author{Divya Thuremella$^{1}$, Yi Yang$^{2,3}$, Simon Wanna$^{2}$, Lars Kunze$^{1,4}$, Daniele De Martini$^{1}$
\thanks{$^{1}$Authors are with the Oxford Robotics Institute, Department of Engineering Science, University of Oxford, UK.}
\thanks{$^{2}$Authors are with the Division of Robotics, Perception, and Learning (RPL), KTH Royal Institute of Technology, Stockholm 114 28, Sweden.
}
\thanks{$^{3}$Authors are with Scania CV AB, Södertälje 151 87, Sweden.}
\thanks{$^{4}$Authors are with the Bristol Robotics Laboratory (BRL), University of the West of England, UK.}
}

\maketitle
\begin{abstract}
This work explores the application of ensemble modeling to the multidimensional regression problem of trajectory prediction for vehicles in urban environments. As newer and bigger state-of-the-art prediction models for autonomous driving continue to emerge, an important open challenge is the problem of how to combine the strengths of these big models without the need for costly re-training. We show how, perhaps surprisingly, combining state-of-the-art deep learning models out-of-the-box (without retraining or fine-tuning) with a simple confidence-weighted average method can enhance the overall prediction. Indeed, while combining trajectory prediction models is not straightforward, this simple approach enhances performance by 10\% over the best prediction model, especially in the long-tailed metrics. We show that this performance improvement holds on both the NuScenes and Argoverse datasets, and that these improvements are made across the dataset distribution. The code~\footnote{https://github.com/dthuremella/Ensemble-of-Pretrained-Models}
for our work is open source.
\end{abstract}

\section{Introduction}

A large number of trajectory prediction models are developed every year~\cite{huangSurveyTrajectoryPredictionMethods2022}, and, in many cases, published models are expensive to retrain (e.g.~\cite{yuanAgentFormerAgentAwareTransformers2021,leeMARTMultiscAleRelational2024a}) or lack published training code (e.g.~\cite{Makansi,dasDecoderonlyFoundationModel2024a,rasulLagLlamaFoundationModels2024}), 
% or where retraining doesn't precisely match the performance of the published model due to stochasticity ~\cite{MART,}, 
This work proposes a straightforward approach to achieve an empirical 10\% improvement in performance with no retraining by running multiple state-of-the-art models and combining their outputs through a confidence-weighted average.

Since one history sequence could yield multiple plausible paths (e.g. a vehicle approaching an intersection could go straight or right), of which the ground truth represents only one out of many potentially likely paths, most vehicle trajectory-prediction methods are multimodal and evaluate their models using the multimodal evaluation metric `Best-of-\textit{N}'.
This metric measures how, among \textit{N} most likely predicted paths, the closest to ground truth resembles the latter \cite{kothariHumanTrajectoryForecasting2021}.
However, this easily becomes impractical, especially for many real-world tasks like path planning and decision-making in autonomous vehicles, since they require precise and convergent trajectory predictions to enable the route planners to avoid potential collision and decide an optimal path \cite{liMinimumofNRethinkingEvaluation2024}. Furthermore, we argue that the `Best-of-N' by itself is not a fair metric, since it allows high scores to be achieved by lucky `shotgun' predictions (evenly spread uninformed guesses) \cite{liMinimumofNRethinkingEvaluation2024,pajouheshgarBackSquareOne2018}. 
Therefore, many works are evaluated using a single prediction, the `most-likely' prediction, and its associated confidence, as this is most easily digestible by downstream tasks in autonomous vehicle systems.

In the scenario of the `most likely' prediction, we show that combining the trajectory outputs of different models improves prediction over the best of the models in the ensemble.
In particular, we achieve a 10\% performance boost on both the NuScenes~\cite{190311027NuScenes} and Argoverse~\cite{wilsonArgoverse2Next2023} datasets by combining the trajectory outputs of an ensemble out of 3 previously state-of-the-art transformer-based models, Autobot \cite{girgis2022latent-autobot}, Wayformer ~\cite{nayakanti2023wayformer}, and MTR ~\cite{shi2022motion}.
These three models differ especially in architecture size -- ranging from 1.5M to 60M parameters -- and training strategies, achieving different performances individually on the prediction task; regardless, the result of the ensemble outperforms the best of the three, even if this entails combining it with models that, on paper, perform worse.

Using this rather simple approach, we see enhancements not only in the overall metrics, but especially in the long-tail metrics -- crucial in safety-critical operations and with imbalanced datasets like those common in trajectory prediction.
Indeed, looking at the errors of the top 1-10\% most difficult examples, we see a consistent improvement across the distribution.
To summarize, in this work, we 1) analyze differences in state-of-the-art trajectory prediction networks and the types of examples they struggle with, and 2) propose a flexible and modular method to combine them into an ensemble to achieve boosted general and long-tail performance at a computational cost. %, and 3) propose a knowledge distillation strategy to train a smaller model to mimic the ensemble's performance, to limited success.

% We will release our code as open source upon acceptance, ensuring full reproducibility.

\begin{figure}
    \centering
    \includegraphics[width=0.55\linewidth]{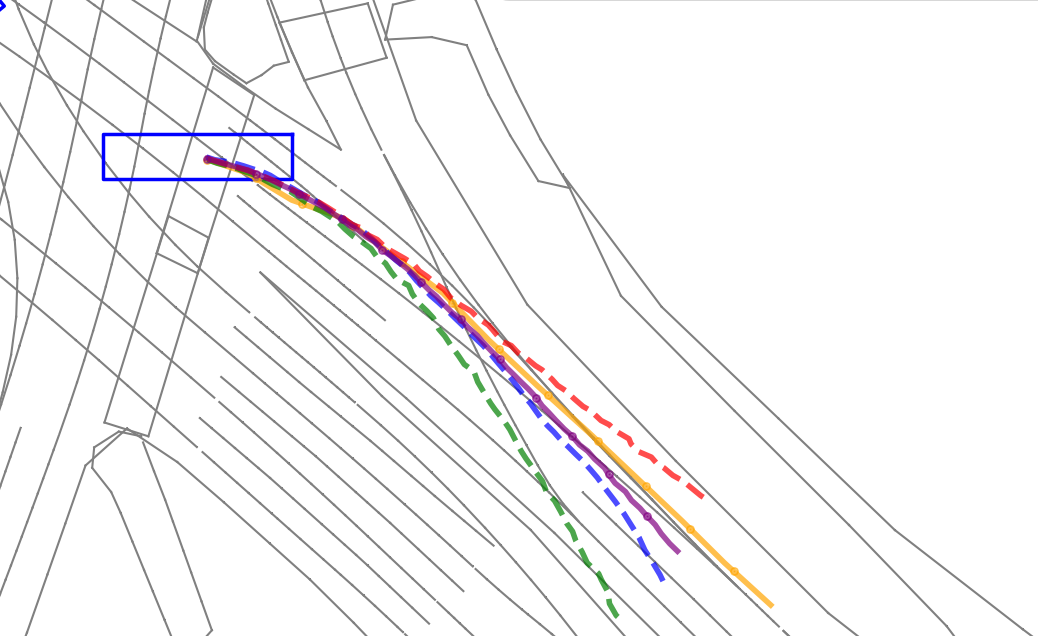}
    \includegraphics[width=0.4\linewidth]{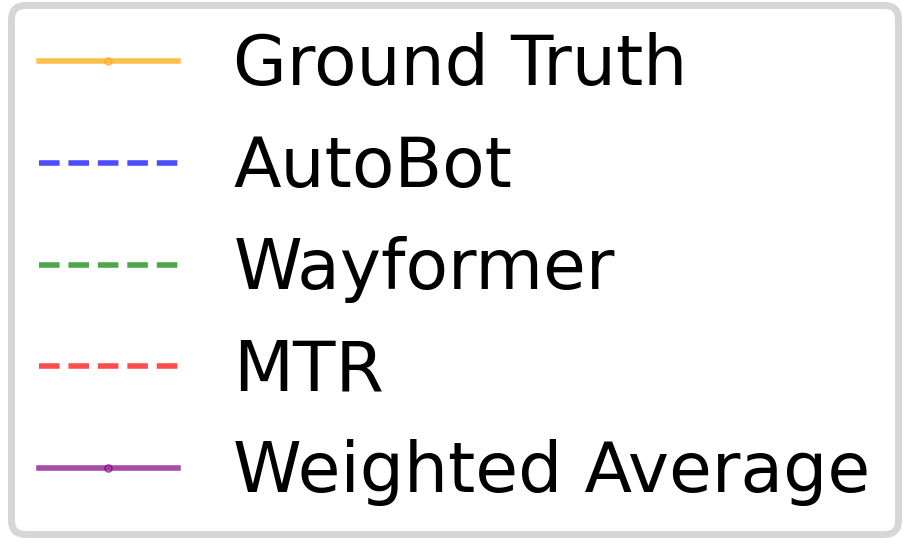}
    \caption{Example of trajectories as predicted by state-of-the-art methods -- AutoBot, Wayformer and MTR -- and our proposed approach. Our approach returns a confidence-weighted average of the three models and outperforms them, as shown here qualitatively.    }
    \label{fig:one_example}
    \vspace{-10pt}
\end{figure}

\section{Related Work}

\subsection{Motion Prediction}
Motion prediction has rapidly advanced with deep learning and large-scale open datasets \cite{Argoverse,caesar2020nuscenes,ettinger2021large_waymo,zhan2019interaction}.
Early raster-based methods utilize convolutional neural networks (CNNs) to extract scene context from bird’s-eye-view images \cite{bansal2018chauffeurnet, chai2019multipath, Hong_2019_CVPR}.
To improve efficiency, recent vector-based approaches represent agents and map elements as structured vectors, leveraging 
networks such as GNNs \cite{gao2020vectornet,liang2020lanegcn, huang2019stgat} and transformers \cite{sceneTransformer, liu2021mmtransformer, zhou2022hivt, huang2022multi,girgis2022latent-autobot} to model traffic interactions.
Separately, query-based decoding strategies inspired by DETR \cite{carion2020end} have been widely adopted to enhance multimodal trajectory generation through flexible, learnable motion queries \cite{shi2024mtr++,zhou2023query}.
In addition, refinement-based methods that first generate coarse trajectory proposals and then refine them have become increasingly popular for improving prediction accuracy \cite{wang2023ganet,wang2023prophnet,zhou2023query,shi2024mtr++,ye2023bootstrap,choi2023r,zhou2024smartrefine}.
More recent works further boost performance by incorporating pre-training strategies with self-supervised learning with improved robustness \cite{yang2023rmp,cheng2023forecast,lansept,zhou2024smartpretrain}.
To unify research efforts, Unitraj \cite{feng2024unitraj} proposes a framework that integrates multiple models and datasets into a single pipeline, facilitating research in this area.
Our work builds upon the Unitraj pipeline.
While new state-of-the-art models continue to emerge, it's important to understand how to effectively combine their strengths. 
% the strengths of multiple state-of-the-art methods simultaneously.

\subsection{Ensemble Modeling in Trajectory Prediction}
Both homogeneous ensemble modeling (where multiple copies of the same architecture are trained in ensemble) and heterogeneous ensemble modeling (where each model in the ensemble has a different architecture) have proven effective in trajectory prediction. \cite{varadarajan2022multipath++} use ensembling at the decoder stage by training a homogenous ensemble of predictor heads. 
% and show that an ensemble of many heads to predict a few trajectories each is more effective than an ensemble of a few heads predicting many trajectories each. 
Meanwhile, ~\cite{nayakPedestrianTrajectoryForecasting2024} and ~\cite{distelzweigStochasticityMotionInformationTheoretic2025} take advantage of the deep ensemble method developed by ~\cite{lakshminarayananSimpleScalablePredictive2017}, which showed that using an ensemble of homogeneous neural networks trained in parallel on the entire training set gave smooth predictions and lower error in the task of object detection.  ~\cite{nayakPedestrianTrajectoryForecasting2024} show that deep ensembles are effective in trajectory prediction by using a homogeneous ensemble of neural networks to improve accuracy and uncertainty estimation, while ~\cite{distelzweigStochasticityMotionInformationTheoretic2025} compare homogeneous deep ensemble configurations to heterogeneous deep ensemble configurations (using 3 heterogenous models: a transformer model~\cite{liuLAformerTrajectoryPrediction2023}, a graph neural network model~\cite{deo2021multimodal}, and an LSTM-based model ~\cite{kimLaPredLaneAwarePrediction2021}) to show that heterogenous ensembles have better model diversity and more accurate predictions and confidences. While ~\cite{nayakPedestrianTrajectoryForecasting2024} and ~\cite{distelzweigStochasticityMotionInformationTheoretic2025} aggregate ensemble model predictions by creating a multimodal Gaussian Mixture Model (GMM) out of the combination of all predictions from the ensemble, and sample this distribution to produce predictions, ~\cite{liEnsembleLearningFramework2022} aggregate their homogeneous ensemble models' predictions through a combination of averaging and majority voting. 
Finally, ~\cite{kimHybridApproachVehicle2021}, the most similar work to ours, also uses a heterogeneous ensemble (made up of a physics-based kalman filter model, a Gauss-Markov process based model, and a deep learning model) and aggregate predictions with a combination of weighted averaging and thresholding. 
% and show that even when using an ensemble of models which have very different accuracies (with the deep-learning model much more accurate on average than the kalman filter), the aggregated (thresholded weighted average) output of the ensemble improves performance beyond what can be achieved by just using the deep learning model. 

% TODO cut??
% Of note, ~\cite{filosCanAutonomousVehicles2020a}, which uses a homogeneous ensemble of neural density estimators, shows that ensembles can more effectively detect and recover from shifts in the dataset distribution by reducing overconfident and catastrophic extrapolation in out-of-distribution scenes. This advantage of ensembles is particularly helpful for the problem of long-tailed learning, due to which performance of examples in under-sampled parts of the dataset distribution suffer, and we confirm these findings through our long-tail metrics.

\subsection{Long-Tailed Trajectory Prediction}

To improve long-tailed performance within trajectory prediction, ~\cite{wangFENDFutureEnhanced2023} and ~\cite{mercuriusAMENDMixtureExperts2024}  use ensemble modeling, while ~\cite{zhangTrACTTrainingDynamics2024} ~\cite{Makansi} and ~\cite{Kozerawski} use regularization techniques. ~\cite{Makansi} and ~\cite{zhangTrACTTrainingDynamics2024}  use contrastive loss on implicit classes of trajectories, while ~\cite{Kozerawski} re-weight the loss according to how rare the trajectory using dataset statistics to determine the measure of rareness. Combining ensemble modeling and long-tailed learning in trajectory prediction, ~\cite{wangFENDFutureEnhanced2023} and ~\cite{mercuriusAMENDMixtureExperts2024} use a two-stage prediction scheme from which ensemble modeling can be done based on the results of the first stage. ~\cite{wangFENDFutureEnhanced2023}, the state of the art, uses a contrastive autoencoder network to separate the dataset into clusters, which are then separated into well-performing and poorly-performing groups, each of which are fed to a different ensemble model. ~\cite{mercuriusAMENDMixtureExperts2024} does not beat the state of the art, but provides a two-stage framework as well from which ensemble modeling can be done no matter the underlying architecture.

To measure long-tailed performance, 
~\cite{Makansi,zhangTrACTTrainingDynamics2024,wangFENDFutureEnhanced2023} use a `difficulty scoring' (based on how easy it is for a Kalman filter ~\cite{Makansi} or the baseline model ~\cite{zhangTrACTTrainingDynamics2024,wangFENDFutureEnhanced2023} to predict the future trajectory) to get the ADE/FDE of the most `difficult' 1, 2, 3, etc. percent of examples.
% , while ~\cite{Kozerawski} calculate the 95th, 98th, and 99th percentile of the distribution of errors to measure long-tail performance. This percentile is equivalent to measuring the CVaR (probability of predictions below a certain error). 
% In our work, we use the same metric used by the state-of-the-art models ~\cite{zhangTrACTTrainingDynamics2024,wangFENDFutureEnhanced2023}, where 
Following them, 
our `difficulty score' is based on each model's own error distribution instead of that of the baseline model, since we have more than one baseline model to compare to.
For more details, see Section \ref{sec:metrics}.

\section{Problem Formulation}
\label{AA}

% Consider a scenario including $N$ agents' trajectories $y$, and their corresponding feature embeddings $A$ over $T$ timestamps, denoted as $A_i \in \mathbb{R} ^{T\times D_{agent}}$, where $i \in [1, N]$ and $D_{agent}$ represents the feature dimension of the agent embedding, which can contain, dependent on the dataset, also yaw angle, velocity and agent size.
% Let then $Map \in \mathbb{R} ^{S \times P \times D_{road}}$ be the surrounding road topology; here, $S$ represents the number of road segments, $P$ denotes the number of points within a segment, and $D_{road}$ signifies the vector feature dimension that includes position coordinates $x,y$ and the validity mask.

% In the context of motion prediction, we are provided with the map $Map$ and historical trajectory $A_{history} \in \mathbb{R} ^{T_{obs}\times D_{agent}}$, where $T_{obs}$ signifies the observed historical timestamps, and our task is to predict the future trajectory $A_{future} \in \mathbb{R} ^{T_{fut}\times D_{agent}}$.

% \yy{
% We consider the task of multi-agent motion prediction over a future horizon.
Given $N$ agents and $T_{\text{history}}$ observed timesteps,
the historical trajectories are denoted as $A_{\text{history}} \in \mathbb{R}^{N \times T_{\text{history}} \times D_{\text{agent}}}$.
$D_{\text{agent}}$ denotes feature dimension, which may include, depending on the dataset, attributes such as position, velocity, and yaw angle.
Let Map $\mathcal{M} \in \mathbb{R} ^{S \times P \times D_{\text{road}}}$ be the surrounding road topology, where $S$ denotes the number of road segments, $P$ the number of points per segment, and $D_{\text{road}}$ the semantic feature dimension. The goal is to predict the future trajectory features $A_{\text{pred}} \in \mathbb{R}^{N \times T_{\text{future}} \times D_{\text{agent}}}$, where $T_{\text{future}}$ is the number of future timesteps.
The predicted agent 2D positions are denoted as $y \in \mathbb{R}^{N \times T_{\text{future}} \times 2}$.
Modern motion prediction models typically output a set of $K$ candidate future trajectories $\{ y^k \}_{k=1}^K$, each with an associated confidence score $c^k$, reflecting the model's certainty for each candidate.

% }
% \section{Problem Formulation}
% \label{AA}

% Consider a scenario including $N$ agents' trajectories $y$, and their corresponding feature embeddings $A$ over $T$ timestamps, denoted as $A_i \in \mathbb{R} ^{T\times D_{agent}}$, where $i \in [1, N]$ and $D_{agent}$ represents the feature dimension of the agent embedding, which can contain, dependent on the dataset, also yaw angle, velocity and agent size.
% Let then $Map \in \mathbb{R} ^{S \times P \times D_{road}}$ be the surrounding road topology; here, $S$ represents the number of road segments, $P$ denotes the number of points within a segment, and $D_{road}$ signifies the vector feature dimension that includes position coordinates $x,y$ and the validity mask.

% In the context of motion prediction, we are provided with the map $Map$ and historical trajectory $A_{history} \in \mathbb{R} ^{T_{obs}\times D_{agent}}$, where $T_{obs}$ signifies the observed historical timestamps, and our task is to predict the future trajectory $A_{future} \in \mathbb{R} ^{T_{fut}\times D_{agent}}$.

\section{Methodology}
% To facilitate the standardization of datasets and metrics across models, we employ UniTraj~\cite{feng2024unitraj}, which standardizes dataset formats and training pipelines for 3 state-of-the-art models, AutoBot~\cite{girgis2022latent-autobot}, Wayformer~\cite{nayakanti2023wayformer}, and MTR~\cite{shi2022motion}, such that they can be easily tested on different datasets such as NuScenes~\cite{190311027NuScenes} and Argoverse~\cite{wilsonArgoverse2Next2023} to perform multi-dataset experiments and ablation studies.
% We use this framework to run pre-trained AutoBot, Wayformer, and MTR models as an ensemble on the NuScenes and Argoverse datasets, and aggregate the ensembled predictions using a confidence-weighted average. 

We use the UniTraj framework~\cite{feng2024unitraj}, which standardizes dataset formats and model training pipelines, facilitating training and evaluation of 3 state-of-the-art models, AutoBot~\cite{girgis2022latent-autobot}, Wayformer~\cite{nayakanti2023wayformer}, and MTR~\cite{shi2022motion}, on different datasets such as NuScenes~\cite{190311027NuScenes} and Argoverse~\cite{wilsonArgoverse2Next2023}. 
We run the pre-trained models as an ensemble and aggregate their predictions using a confidence-weighted average.

\subsection{Base Models}
Autobot ~\cite{girgis2022latent-autobot}, Wayformer ~\cite{nayakanti2023wayformer}, and MTR ~\cite{shi2022motion} have similar transformer-based encoder and decoder architectures and predict a gaussian mixture model (GMM) probability distribution for each future timestep, but differ in size -- AutoBot, the smallest, uses only 1.5M parameters, Wayformer uses 16.5M parameters, and MTR uses 60.1M parameters -- and employ different loss strategies.
% : while AutoBot uses the negative log-likelihood (NLL) loss used in previous benchmark trajectory prediction works (e.g. ~\cite{salzmannTrajectronDynamicallyFeasibleTrajectory2021}), which aims to increase the likelihood of the ground truth given the predicted distribution, Wayformer uses the NLL of the predicted mode closest to the ground truth given the predicted distribution, as well as an additional classification loss. MTR, on the other hand, uses the NLL of the ground truth over the chosen gaussian mode, calculated at each decoder layer, and adds a loss term which performs the auxiliary task of predicting future interactions between agents.
% The three models also differ in size, with . 
% 
Here, we briefly describe the models included in the ensemble (AutoBot, Wayformer, and MTR) in more detail.

\subsubsection{AutoBot}
Autobot \cite{girgis2022latent-autobot} is a transformer-based model that learns equivariant representations by modeling the joint distribution of agent trajectories.
At each timestep, AutoBot predicts the parameters of a bivariate Gaussian distribution — means $(\mu_x, \mu_y)$, standard deviations $(\sigma_x, \sigma_y)$, and correlation coefficient $\rho$ — for the future position of each agent.  
The model is trained by minimizing the negative log-likelihood (NLL) of the ground-truth future positions under the predicted distribution:

\begin{equation}
\mathcal{L}_{\text{NLL}} = -\sum_{t=1}^{T} \log p\left( y^t_x, y^t_y \mid \mu_x^t, \mu_y^t, \sigma_x^t, \sigma_y^t, \rho^t \right),
\end{equation}

where $p(\cdot)$ is the probability density function of a bivariate Gaussian.
\subsubsection{Wayformer}
Wayformer \cite{nayakanti2023wayformer} extends transformer architectures with factorized attention and learnable latent queries, followed by an early fusion of multimodal inputs.
At each layer, the input representations 
$\mathbf{z}$ are updated based on agent tokens, lane tokens, and waypoint tokens separately before fusion using attention mechanism via:
\begin{equation}
\mathbf{z}' = \text{Fusion}(\mathrm{Attn}_{\text{agent}}(\mathbf{z}),\ \mathrm{Attn}_{\text{lane}}(\mathbf{z}),\ \mathrm{Attn}_{\text{wp}}(\mathbf{z})),
\end{equation}
where $\mathbf{z}$ denotes the input features and $\mathbf{z}'$ the updated features after multi-axis attention and fusion.

Wayformer predicts future trajectories by outputting a multi-modal gaussian (GMM) probability distribution, and trains by using an NLL loss which maximizes the likelihood of the gaussian mode closest to the ground truth under the predicted distribution, as well as a classification loss on mode selection which encourages the model to choose the most accurate mode. 

% loss on displacement errors combined with a classification loss .

\subsubsection{MTR}
MTR (Motion Transformer Refinement) \cite{shi2022motion} extends the transformer architecture by using a set of learnable motion query pairs, where the first element in the pair learns spatially distributed intention points to direct coarse intention prediction for a specific mode of the output, and the other produces trajectory refinements for those points. Furthermore, they introduce an auxiliary task of predicting future interactions between agents to guide the training.

% a two-stage framework that first generates coarse trajectory hypotheses and then refines them through localized motion queries.  
% In the first stage, a small set of global queries $\{q_{\text{global}}\}$ is used to generate coarse trajectories $\{\hat{y}_{\text{coarse}}\}$.  
% In the second stage, refinement queries $\{q_{\text{local}}\}$ are conditioned on the coarse outputs to predict refined trajectories $\{\hat{y}_{\text{refine}}\}$.  

% The training objective jointly optimizes the predicted GMM distribution and the auxiliary task. The GMM distribution is optimized via an NLL loss (the negative log-likelihood of the ground-truth under the chosen mode of the predicted GMM), calculated at each decoder layer and summed up. Meanwhile, the auxiliary loss is defined as

The training objective jointly optimizes the predicted GMM distribution and the auxiliary task:
\begin{equation}
\mathcal{L} = \mathcal{L}_{\text{auxiliary}} + \mathcal{L}_{\text{NLL}},
\end{equation}
% 
% where
\begin{equation}
\mathcal{L}_{\text{auxiliary}} = \text{L1}(A_{\text{pred}}, A_{\text{gt}})
\end{equation}

% \begin{equation}
% \mathcal{L}_{\text{refine}} = \sum_{k=1}^{K} \text{SmoothL1}(\hat{y}^k_{\text{refine}}, y_{\text{gt}})
% \end{equation}

Here, $\mathcal{L}_{\text{NLL}}$ is the negative log-likelihood of the ground-truth under the chosen mode of the predicted GMM, calculated at each decoder layer and summed up; $A_{\text{pred}}$ is the predicted future positions and velocities of the agents, while $A_{\text{gt}}$ represents the corresponding ground truth values. By combining coarse global intention with fine-grained local refinement query pairs, MTR achieves strong trajectory prediction performance.

\subsection{Confidence-Weighted Model Ensemble}
% \subsubsection{Ensemble Prediction}
We interpret model ensembling as a form of uncertainty-aware decision fusion.
Given three trajectory prediction models—AutoBot (AB), Wayformer (WF), and MTR—we assume each model provides a future trajectory along with an internal confidence score that reflects its estimated certainty.
% Let $\hat{y}_{\text{AB}}, \hat{y}_{\text{WF}}, \hat{y}_{\text{MTR}}$ denote the most confident trajectory from each model, and $c_{\text{AB}}, c_{\text{WF}}, c_{\text{MTR}}$ be their corresponding confidence scores.  
% To fully leverage the strengths of AutoBot (AB), Wayformer (WF), and MTR, we adopt a confidence-weighted ensemble strategy.  
For each sample, we extract the most confident trajectory $\hat{y}_{\text{AB}}$, $\hat{y}_{\text{WF}}$, and $\hat{y}_{\text{MTR}}$ from the three models, along with their associated confidence scores $c_{\text{AB}}$, $c_{\text{WF}}$, and $c_{\text{MTR}}$.  
Since all models predict a GMM distribution of trajectories, assigning a confidence to each predicted trajectory mode and training these confidences using a form of the NLL loss, the confidences are comparable between models.

First, the confidence scores are normalized to sum to one:
\begin{equation}
\tilde{c}_i = \frac{c_i}{\sum_j c_j}, \quad i \in \{\text{AB}, \text{WF}, \text{MTR}\},
\end{equation}

The final predicted trajectory $\hat{y}_{\text{final}}$ is then obtained as a weighted average over the three models:
\begin{equation}
\hat{y}_{\text{final}} = \sum_{i} \tilde{c}_i \cdot \hat{y}_i, \quad \text{for } i \in \{\text{AB}, \text{WF}, \text{MTR}\}.
\end{equation}
This formulation can be viewed as minimizing the expected trajectory error under a model-uncertainty prior, where higher-confidence models are trusted more.  
It also mitigates the impact of failure cases in any individual model by adaptively redistributing trust across models, leading to more robust final predictions.

% \subsubsection{Confidence of Ensemble Prediction}
To calculate how confident the ensemble model is in its weighted-average prediction, we take inspiration from ~\cite{nayakPedestrianTrajectoryForecasting2024, distelzweigStochasticityMotionInformationTheoretic2025,lakshminarayananSimpleScalablePredictive2017}, and use the weighted variance across the three models as an inverse measure of confidence.

We first calculate the 2D covariance matrix of the ensemble's prediction, given by:
\begin{equation}
\sigma^2_{\text{final}} = \sum_i \tilde{c}_i \left(  (\hat{y}_i - \hat{y}_{\text{final}})(\hat{y}_i - \hat{y}_{\text{final}})^T \right).
\end{equation}
where the determinant of this covariance matrix, $\det(\sigma^2_{\text{final}})$, captures the overall dispersion of the predictions in 2D space. A higher determinant indicates greater disagreement among models, while a lower determinant suggests more consistent predictions.
To transform this measure into a confidence score $c_{\text{final}}$, we apply an inverse transformation defined as
\begin{equation}
c_{\text{final}} = \frac{1}{1 + \det(\sigma^2_{\text{final}})},
\end{equation}
which ensures the confidence is bounded between 0 and 1.

% where the inverse of $\sigma^2_{\text{final}}$ is viewed as a measure of the confidence of our model. 

% \subsection{Knowledge Distillation on the Ensemble}
% \input{tables/kd2}

% The biggest drawback of using the proposed ensemble method is that the ensemble gives better performance at the computational cost of having to run multiple models at inference time. Therefore, we explore using a response-based knowledge distillation model ~\cite{gouKnowledgeDistillationSurvey2021} to see if we can train one model to achieve the same output that the ensemble can achieve. In response-based knowledge distillation, the student model (the smaller model being trained) is trying to mimic the final prediction of the teacher model (the ensemble), and is a method which has shown to be effective for model compression ~\cite{gouKnowledgeDistillationSurvey2021}. 

% In our work, we use the AutoBot architecture as the student model, since is is the smallest and therefore cheapest to train, and train it to predict the outputs of the ensemble of three pre-trained models instead of the ground truth. As shown in Table ~\ref{tab:kd}, while this method did not match the performance of the ensemble, or beat the results of the larger, more accurate Wayformer and MTR models, it led to some small improvements in most-likely performance over the original pre-trained Autobot model. 

\section{Experimental Setup}

\subsection{Datasets}

We show our results on two different datasets, NuScenes \cite{190311027NuScenes} and Argoverse2 \cite{wilsonArgoverse2Next2023}. While both datasets are in urban driving environments and taken from a moving vehicle, NuScenes includes data from both left-side driving and right-side driving cities, and Argoverse2 covers more cities and contains more data.

\subsubsection{NuScenes}
The NuScenes dataset ~\cite{190311027NuScenes} consists of 1000 scenes with 5.5 hours of footage labeled at 2Hz. It has 32k trajectories taken from a moving vehicle in 4 neighborhoods within Boston and Singapore, and includes HD semantic maps with 11 annotated layers, including pedestrian crossings, walkways, stop lines, traffic lights, road dividers, lane dividers, and driveable areas ~\cite{feng2024unitraj,salzmannTrajectronDynamicallyFeasibleTrajectory2021}. 
\subsubsection{Argoverse2}
The Argoverse 2 dataset~\cite{wilsonArgoverse2Next2023}, includes 6 million lidar frames across 20,000 scenes with 2k hours of footage. It has almost 180k trajectories taken in 6 cities of different sizes across the US: Miami, Pittsburgh, Austin, Washington DC, Dearborn, and Palo Alto~\cite{wilsonArgoverse2Next2023,feng2024unitraj}. HD semantic maps in the Argoverse 2 dataset include lanes, lane boundaries (dashed white, dashed yellow, double yellow, etc), and crosswalks.

\subsection{Metrics}
\label{sec:metrics}
In this section we describe the metrics upon which we evaluate our ensemble. Specifically, we use the most-likely metric, as described below, and report our long-tailed performance. Although Autobot~\cite{girgis2022latent-autobot}, Wayformer~\cite{nayakanti2023wayformer} and MTR~\cite{shi2022motion} report 'Best-of-N' metrics like minADE, minFDE, and miss rate, these metrics are specific to models that predict multiple possible trajectories for each example, and since our ensemble model only outputs one prediction, we cannot report these metrics.

% Most methods that use the NuScenes and Argoverse datasets, to our knowledge, use the best-of-20 metric but this is an impractical metric to use when considering downstream tasks like path planning in autonomous vehicles ~\cite{liMinimumofNRethinkingEvaluation2024} because it allows high scores to be achieved by lucky 'shot gun' predictions (evenly spread uninformed guesses) ~\cite{liMinimumofNRethinkingEvaluation2024,pajouheshgarBackSquareOne2018}. Therefore, we report the 'most likely' metric, which only outputs one (the most likely) trajectory and its associated confidence, as this is what is most easily digestible by downstream tasks. 

\subsubsection{The Most-Likely Metric} compares the average distance error (ADE) and final distance error (FDE) between the ground truth path and the single most likely prediction output by the model. Such methods allow applications to plan for only the most likely futures and are therefore more useful in practice. 

% \subsubsection{Long-Tailed Metrics} To show improvement in the long tail, we also evaluate the performance on only the long tail of the dataset. In our work, we use the same metric used by the state-of-the-art models ~\cite{zhangTrACTTrainingDynamics2024,wangFENDFutureEnhanced2023}, where our 'difficulty score' is based on each model's own error distribution instead of that of the baseline model (since we have more than one baseline model to compare to). 

% \yy{
\subsubsection{Long-Tailed Metrics}:
We report long-tailed metrics that focus on a subset of difficult predictions to evaluate performance on challenging scenarios.
% Two variants are commonly used in the literature.
% The first approach~\cite{Makansi,zhangTrACTTrainingDynamics2024,wangFENDFutureEnhanced2023} defines a \textit{difficulty score}—based on the error made by a baseline model (e.g., Kalman filter or learned predictor)—and computes the average ADE/FDE over the most difficult 1\%, 2\%, or 3\% subset of the data.  
% The second approach~\cite{Kozerawski} instead reports the 95th, 98th, and 99th percentiles of the error distribution, effectively measuring the Conditional Value at Risk (CVaR), which captures the threshold below which a given percentage of the worst cases lie.
% In our work, we adopt the former strategy.  
Specifically, we compute the per-sample error for each model, sort the examples accordingly, and report the average error over the top $K\%$ subset based on each model’s own error distribution.  
Formally, let $\mathcal{D} = \{e_1, e_2, \dots, e_N\}$ be the set of per-sample errors (ADE or FDE), and let $\mathcal{D}_{\text{top-}K\%}$ be the subset with the highest $K\%$ errors. We define the Top-$K\%$ error as:

\begin{equation}
\text{Top-}K\%\text{~~Error} = \frac{1}{|\mathcal{D}_{\text{top-}K\%}|} \sum_{e_i \in \mathcal{D}_{\text{top-}K\%}} e_i
\end{equation}
% }

\section{Results and Discussion}

\subsection{Analysis of Base Models}

To help motivate the analysis of our results, we first analyze the strengths and weaknesses of the 3 base models. 

\begin{figure*}[!h]
    \vspace{5pt}
    \centering
    \includegraphics[height=0.2\linewidth]{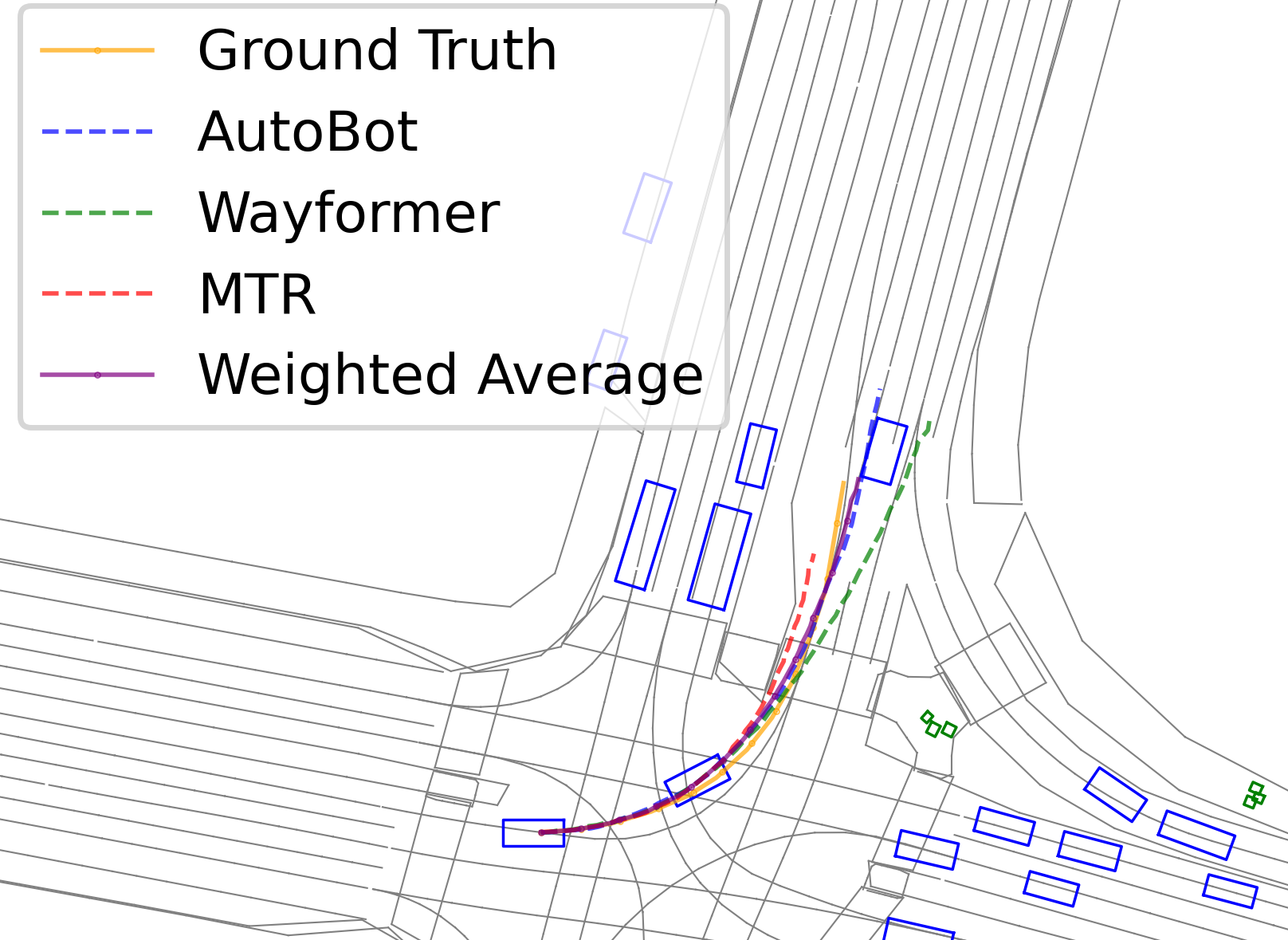}
    \includegraphics[height=0.2\linewidth]{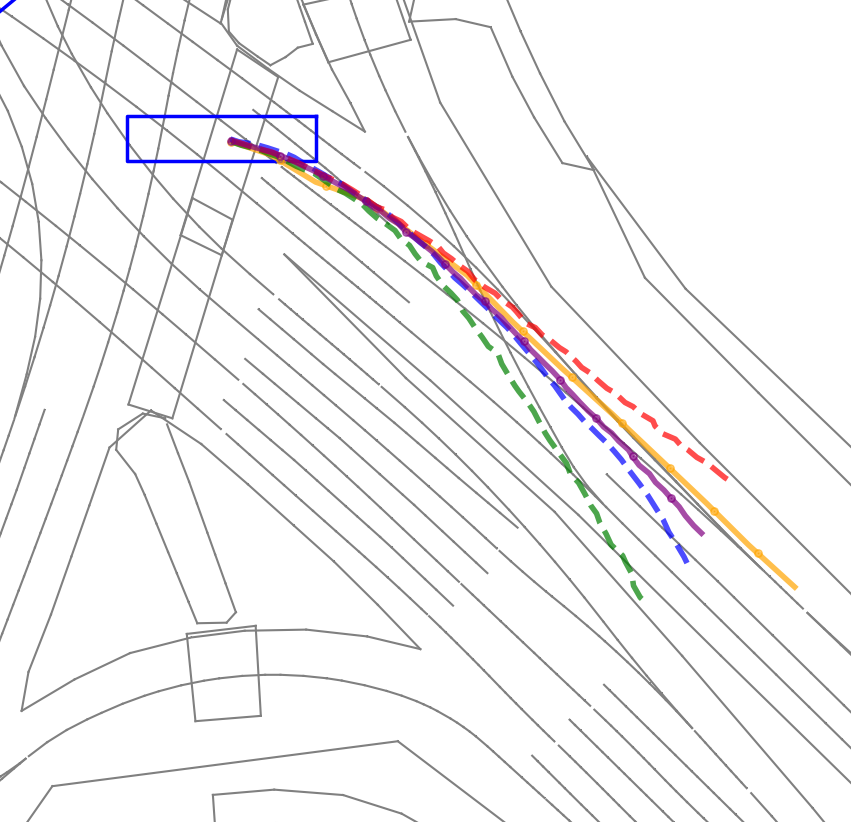}
    \includegraphics[height=0.2\linewidth]{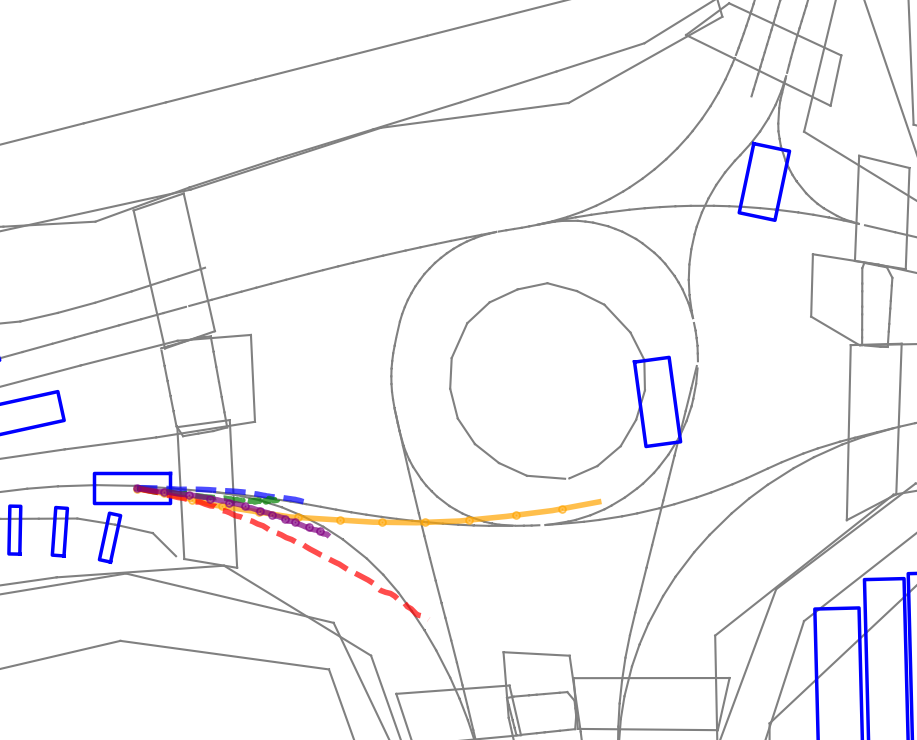}
    \includegraphics[height=0.2\linewidth]{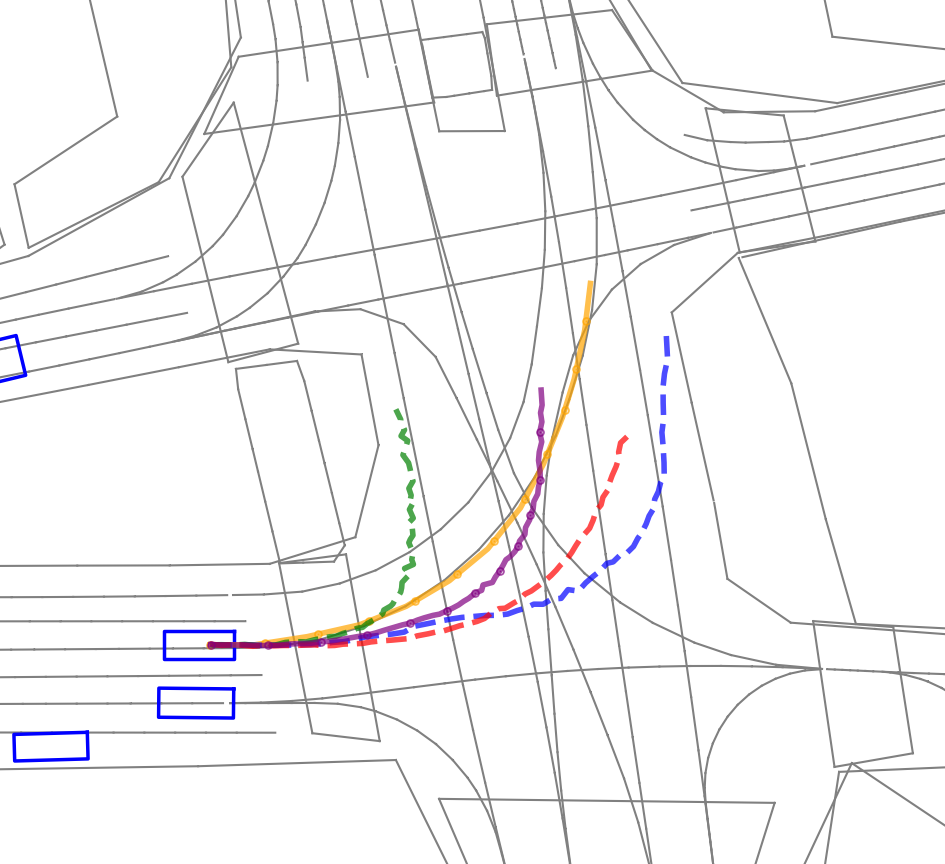}

    \includegraphics[height=0.2\linewidth]{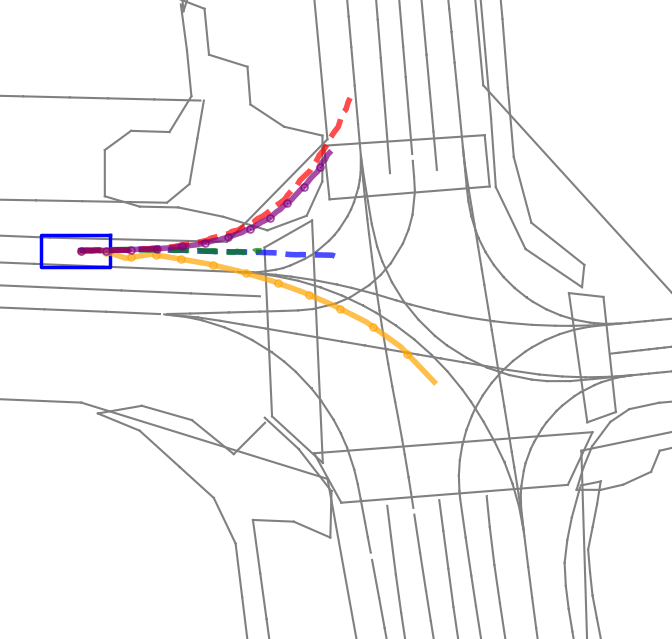}
    \includegraphics[height=0.2\linewidth]{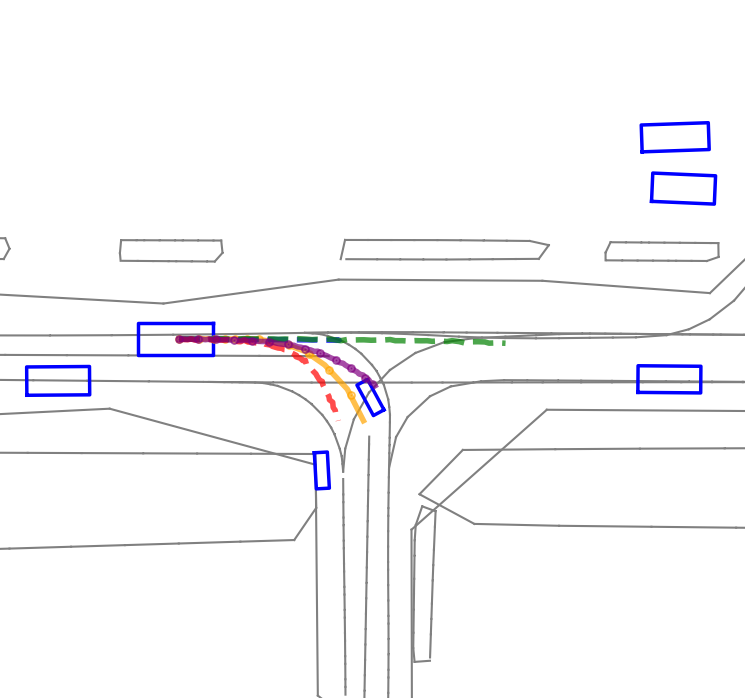}
    \includegraphics[height=0.2\linewidth]{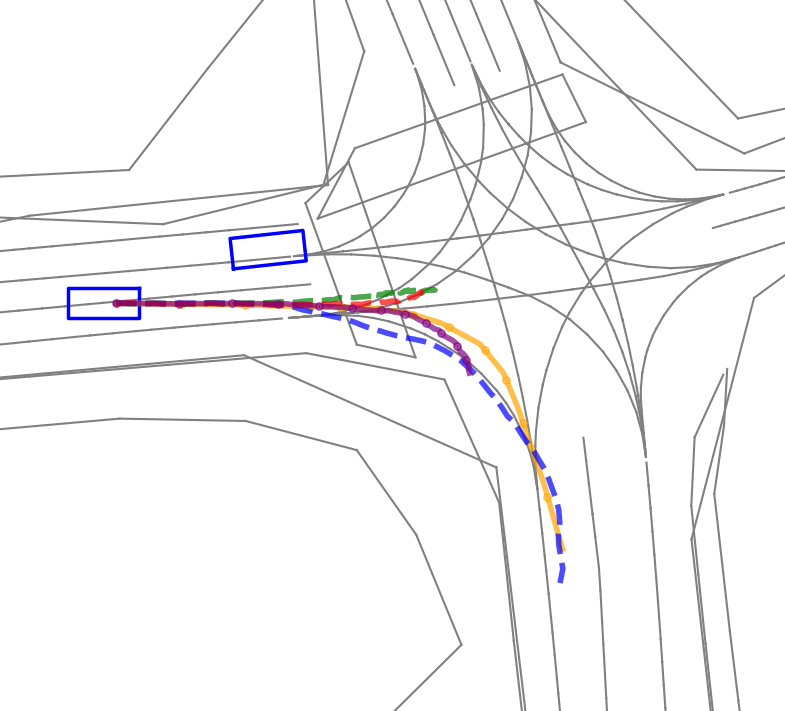}
    \includegraphics[height=0.2\linewidth]{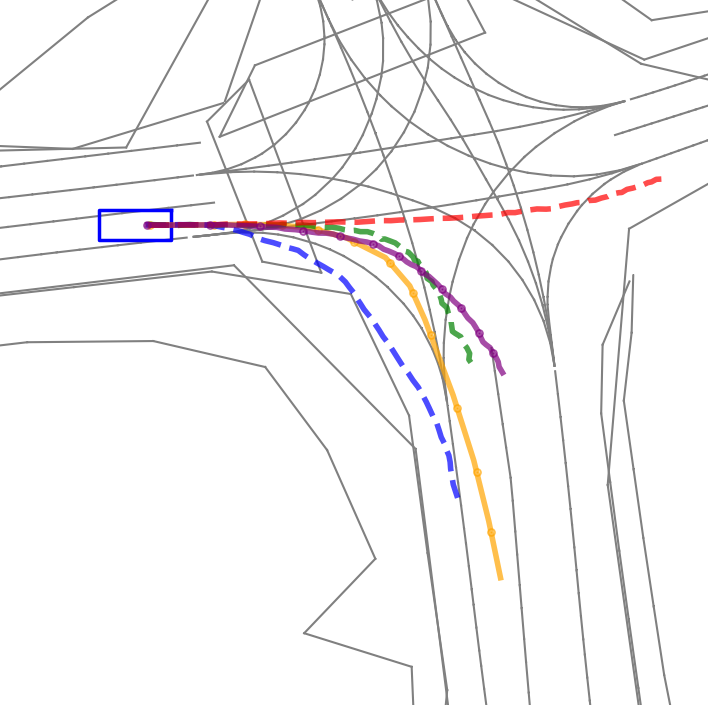}
    \caption{Qualitative Results on NuScenes dataset: the top row contains examples that show improvement while the bottom row demonstrates different types of failure cases. Failure modes with high variance between models are marked as low confidence, as discussed in Section VI-C. The figures show the ground truth, the most-likely prediction for each of the three models (AutoBot, Wayformer, and MTR), and the weighted average of these three predictions.}
    \label{fig:nuScenes}
    \vspace{-15pt}
\end{figure*}

\begin{figure*}[!h]
    \centering
    \includegraphics[height=0.15\linewidth]{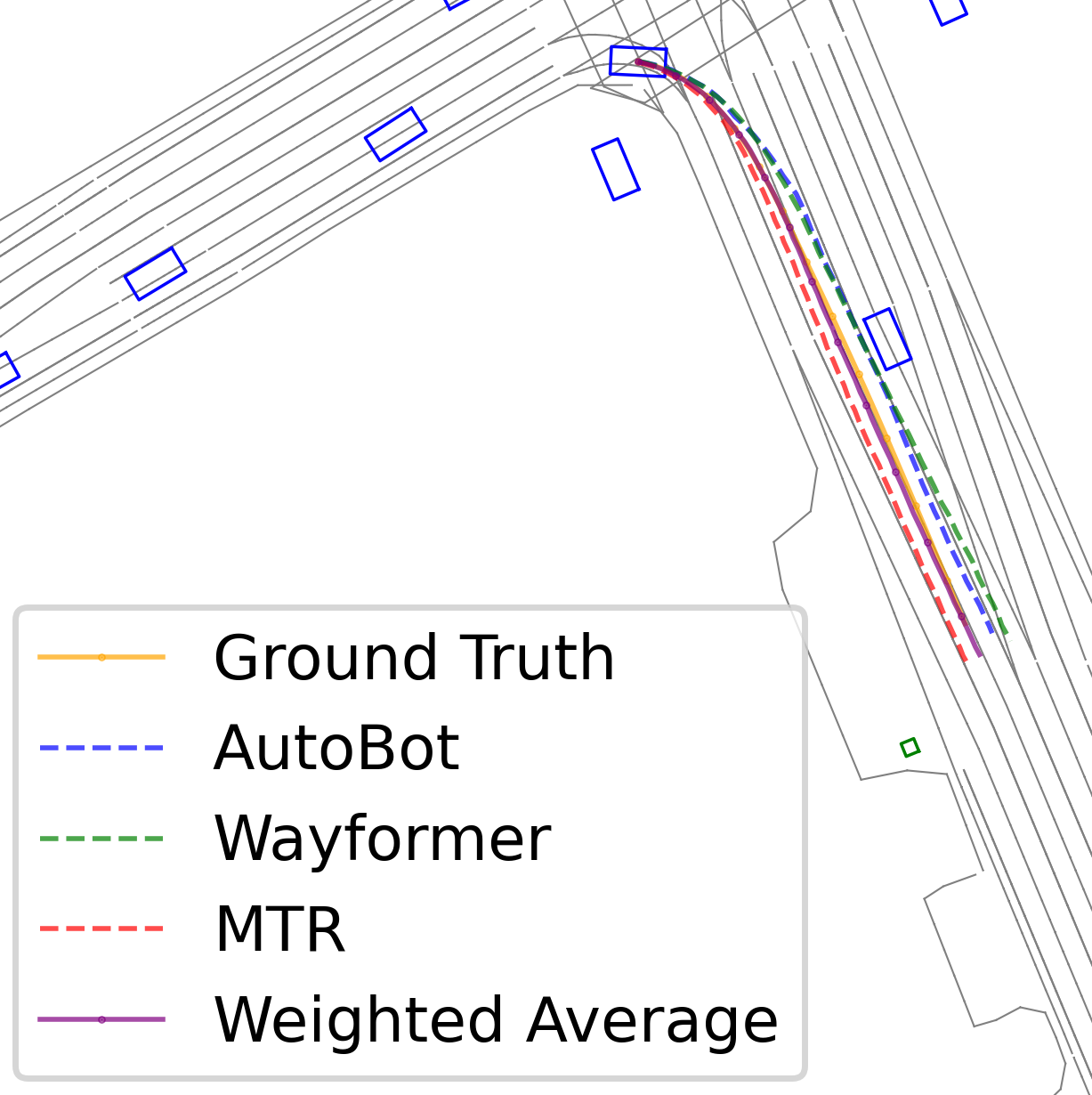}
    \includegraphics[height=0.15\linewidth]{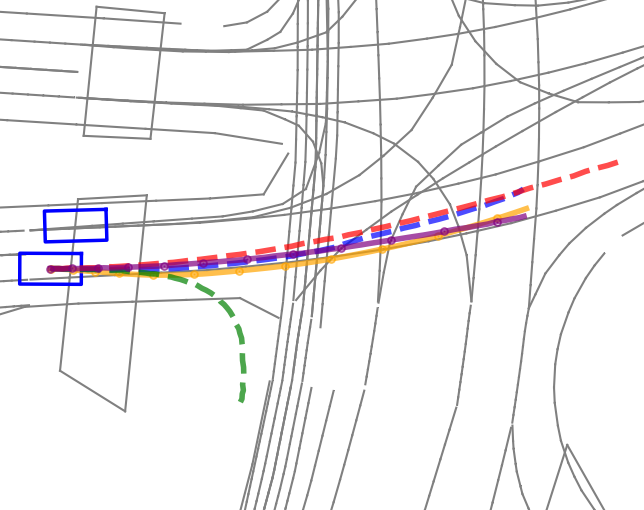}
    \includegraphics[height=0.15\linewidth]{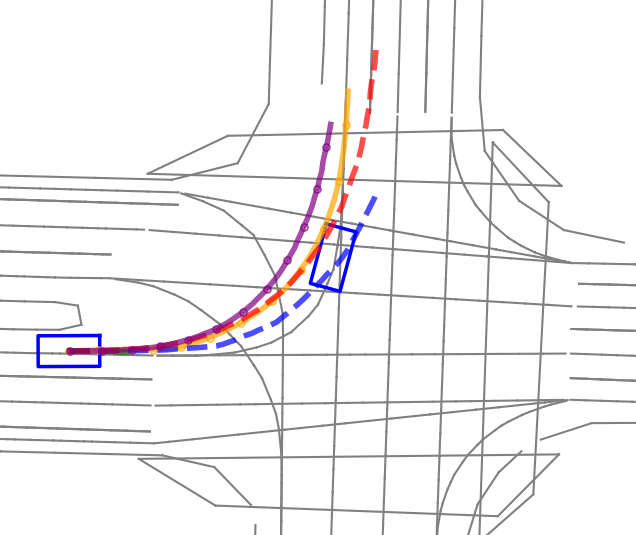}
    \includegraphics[height=0.15\linewidth]{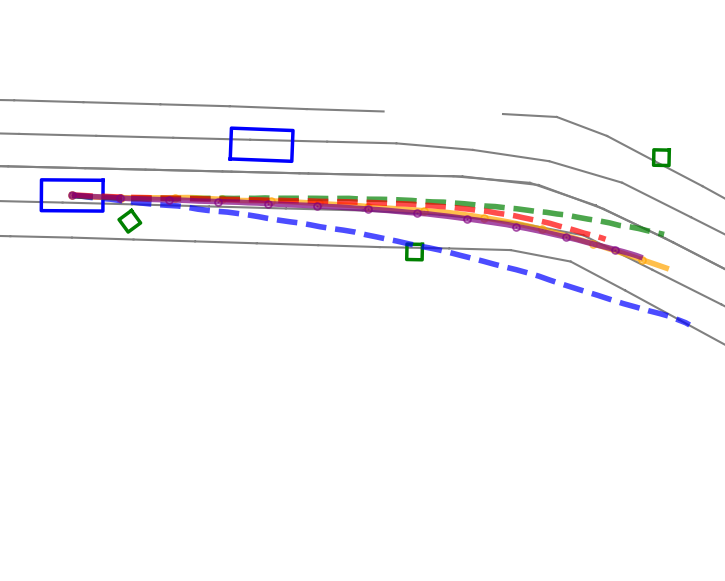}

    \includegraphics[height=0.14\linewidth]{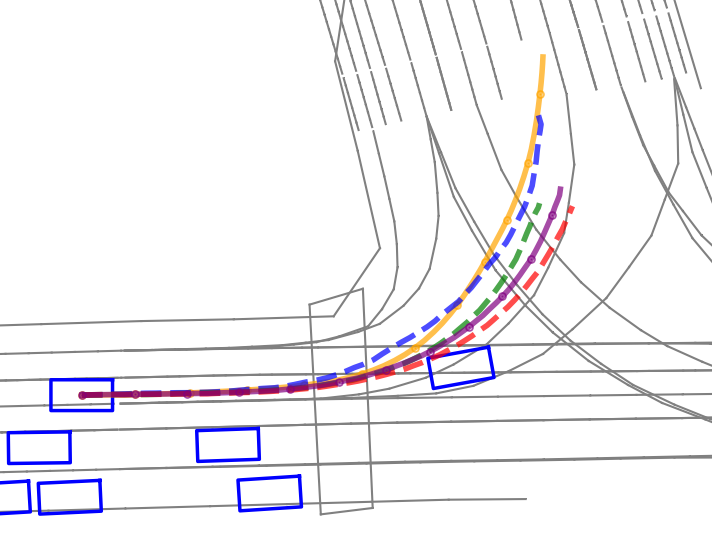}
    \includegraphics[height=0.14\linewidth]{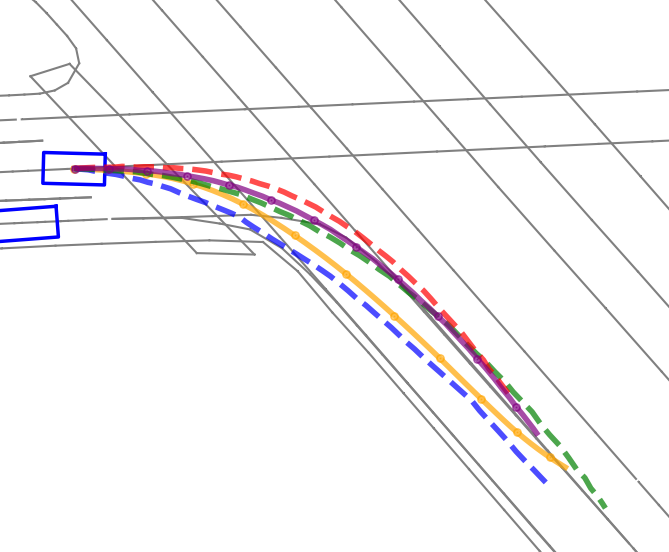}
    \includegraphics[height=0.14\linewidth]{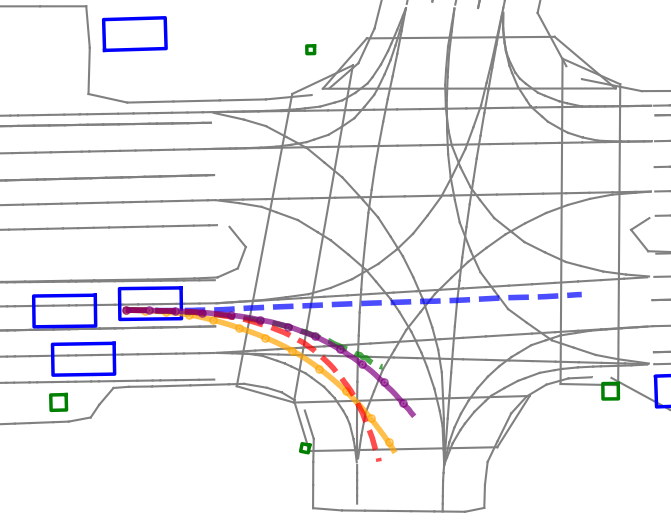}
    \includegraphics[height=0.14\linewidth]{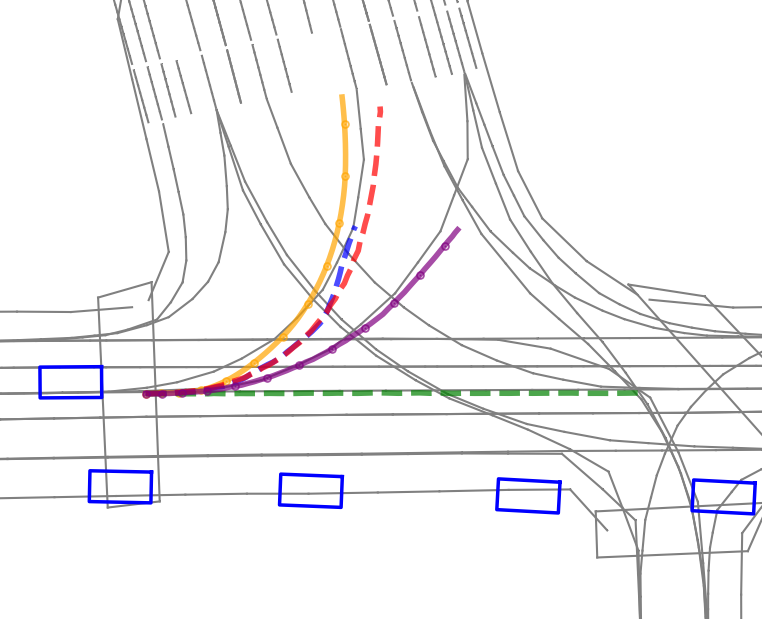}

    \caption{Qualitative Results on Argoverse 2 dataset: the top row contains examples that show improvement while the bottom row demonstrates different types of failure cases. Failure modes with high variance between models are marked as low confidence, as discussed in Section VI-C. The figures show the ground truth, the most-likely prediction for each of the three models (AutoBot, Wayformer, and MTR), and the weighted average of these three predictions.}
    \label{fig:av2}
    \vspace{-10pt}
\end{figure*}

While MTR produces a set of trajectories that are low in variance, where the most likely trajectory is often very accurate, Wayformer produces a set of trajectories that are more widely distributed, where the most likely mode is not always the one closest to the ground truth. This may be influenced by the fact that MTR is 4 times bigger than Wayformer, and its architecture is specifically suited to learn which mode to choose. While both MTR and Wayformer predict trajectories that stick to the lane lines on the map, AutoBot doesn't always follow lane markings; this means that while it's often less correct than the other models, it is better at predicting sharp turns and irregular trajectories that don't perfectly match the neat lane lines, as shown in Figures~\ref{fig:nuScenes} and ~\ref{fig:av2}. Again, this may be because AutoBot is very small (10 times smaller than Wayformer) and unable to achieve the model capacity required to properly parse the map.
However, there are many examples, especially in the Argoverse dataset, where AutoBot beats the other two models, as shown in the bottom left examples in Figure~\ref{fig:av2} and bottom right examples in Figure~\ref{fig:nuScenes}.
Although AutoBot performs much worse than the others on the NuScenes dataset, and both AutoBot and Wayformer perform much worse than MTR on the Argoverse dataset, all three models are still important members of the ensemble and still help improve ensemble performance, as we show in our results.

% small table - how many percent of the scenario is overlapping
% - how many percent of examples are shared in ABC - 

\begin{table*}
\vspace{5pt}
\centering
\begin{tabular}{lccccccc} 
% \midrule
\midrule
Method                 & Top 1\%       & Top 2\%       & Top 3\%       & Top 4\%       & Top 5\%       & Top 10\%     & Overall      \\ 
\midrule
AutoBot (AB)~\cite{girgis2022latent-autobot}           & 15.49 / 38.22 & 13.64 / 34.02 & 12.62 / 31.58 & 11.90 / 29.82 & 11.32 / 28.41 & 9.47 / 23.91 & 3.23 / 8.08  \\
Wayformer (WF)~\cite{nayakanti2023wayformer}         & 13.47 / 36.58 & 11.87 / 32.35 & 10.88 / 29.71 & 10.14 / 27.87 & ~9.56 / 26.42  & 7.80 / 21.72 & 2.49 / 6.77  \\
MTR~\cite{shi2022motion}                    & 13.03 / 35.30 & 11.45 / 31.18 & 10.52 / 28.73 & ~9.84 / 26.94  & ~9.30 / 25.48  & 7.67 / 20.85 & 2.55 / 6.59  \\ 
\midrule
\textit{Ensemble (Weighted) w.} &               &               &               &               &               &              &              \\
AB + WF                & 12.69 / 33.12 & 11.08 / 29.31 & 10.12 / 27.02 & 9.49 / 25.40  & 9.01 / 24.13  & 7.51 / 20.20 & 2.50 / 6.57  \\
AB + MTR               & 11.64 / 30.80 & 10.31 / 27.48 & ~9.48 / 25.33  & 8.89 / 23.79  & 8.43 / 22.58  & 7.00 / 18.81 & 2.40 / 6.20  \\
WF + MTR               & 11.70 / 31.91 & 10.20 / 28.07 & ~9.32 / 25.61  & 8.69 / 23.87  & 8.22 / 22.56  & 6.78 / 18.60 & 2.27 / 5.98  \\
AB + WF + MTR          & 11.27 / 30.30 & ~9.89 / 26.80  & ~9.04 / 24.56  & 8.46 / 22.98  & 8.02 / 21.77  & 6.68 / 18.10 & 2.25 / 5.91  \\
\midrule
\end{tabular}\caption{Most-likely ADE / FDE performance using the NuScenes dataset on the long tail (the Top 1-10\% most difficult examples) as well as over all the data.}
\label{tab:nus}
\end{table*}

\begin{table*}
\centering
\begin{tabular}{lccccccc} 
\midrule
Method                 & Top 1\%       & Top 2\%       & Top 3\%       & Top 4\%       & Top 5\%       & Top 10\%     & Overall      \\ 
\midrule
AutoBot (AB)~\cite{girgis2022latent-autobot}           & 14.60 / 36.85 & 12.74 / 32.63 & 11.62 / 29.97 & 10.83 / 28.09 & 10.22 / 26.57 & 8.31 / 21.75 & 2.46 / 6.27  \\
Wayformer (WF)~\cite{nayakanti2023wayformer}          & 15.10 / 38.31 & 13.21 / 33.94 & 12.11 / 31.25 & 11.31 / 29.26 & 10.67 / 27.72 & 8.67 / 22.64 & 2.50 / 6.36  \\
MTR~\cite{shi2022motion}                    & 13.20 / 34.66 & 11.25 / 29.70 & 10.11 / 26.78 & ~9.30 / 24.67  & ~8.68 / 23.04  & 6.82 / 18.01 & 1.94 / 4.75  \\ 
\midrule
\textit{Ensemble (Weighted) w.} &               &               &               &               &               &              &              \\
AB + WF                & 13.75 / 34.78 & 12.04 / 30.86 & 11.00 / 28.44 & 10.26 / 26.65 & 9.69 / 25.24  & 7.88 / 20.63 & 2.33 / 5.94  \\
AB + MTR               & 12.07 / 31.78 & 10.29 / 27.24 & ~9.26 / 24.52  & ~8.55 / 22.63  & 8.01 / 21.20  & 6.36 / 16.78 & 1.86 / 4.62  \\
WF + MTR               & 12.01 / 31.69 & 10.21 / 27.11 & ~9.18 / 24.40  & ~8.48 / 22.51  & 7.94 / 21.07  & 6.32 / 16.74 & 1.86 / 4.63  \\
AB + WF + MTR          & 11.73 / 30.86 & 10.01 / 26.43 & ~9.05 / 23.91  & ~8.38 / 22.16  & 7.87 / 20.80  & 6.31 / 16.68 & 1.87 / 4.69  \\
\midrule
\end{tabular}
\caption{Most-likely ADE / FDE performance using the Argoverse 2 dataset on the long tail (the Top 1-10\% most difficult examples) as well as over all the data.}
\label{tab:av2}
\end{table*}
\begin{figure}
    \vspace{6pt}
    \centering
    \includegraphics[width=0.4\linewidth]{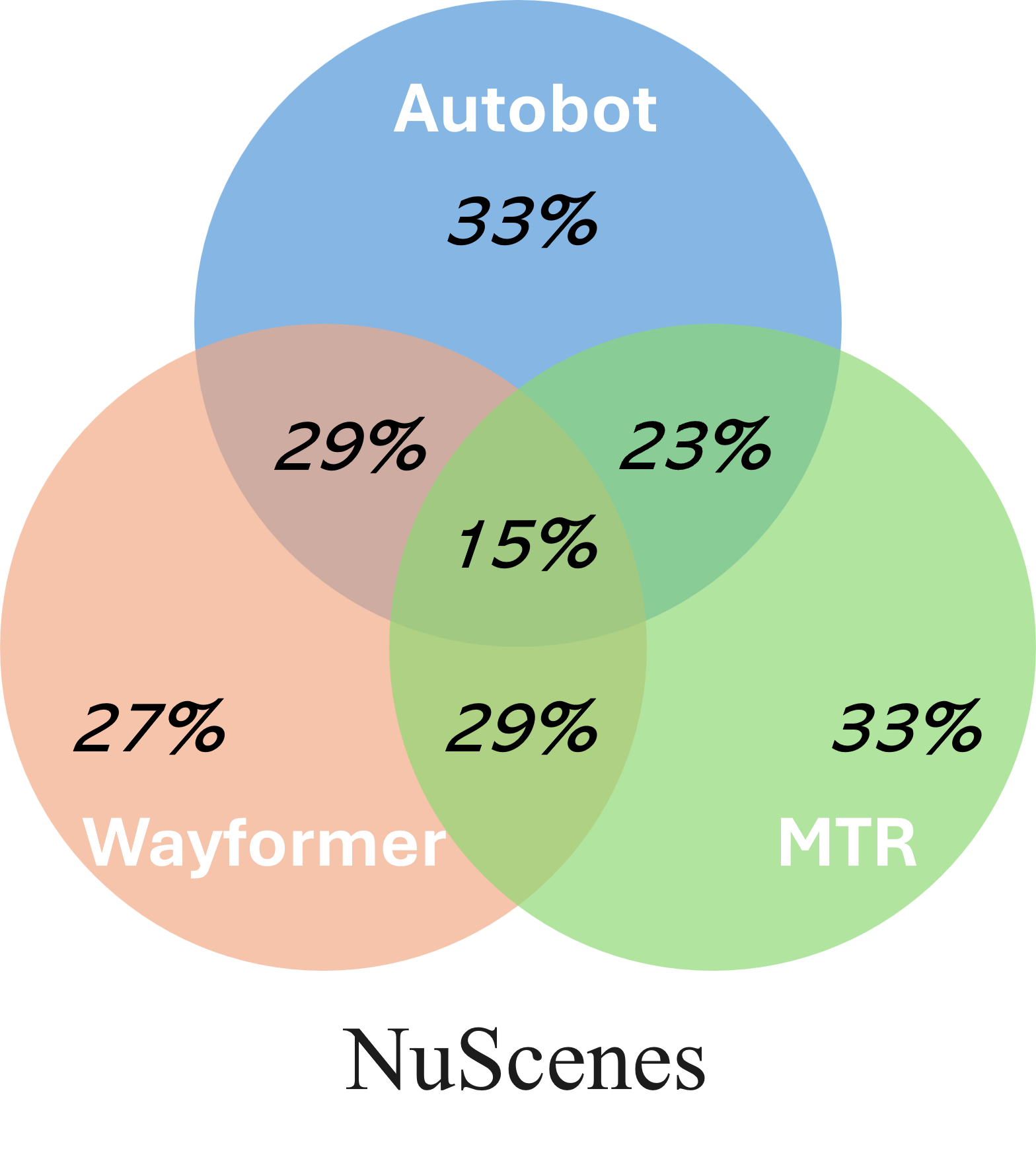}~~~
    \includegraphics[width=0.4\linewidth]{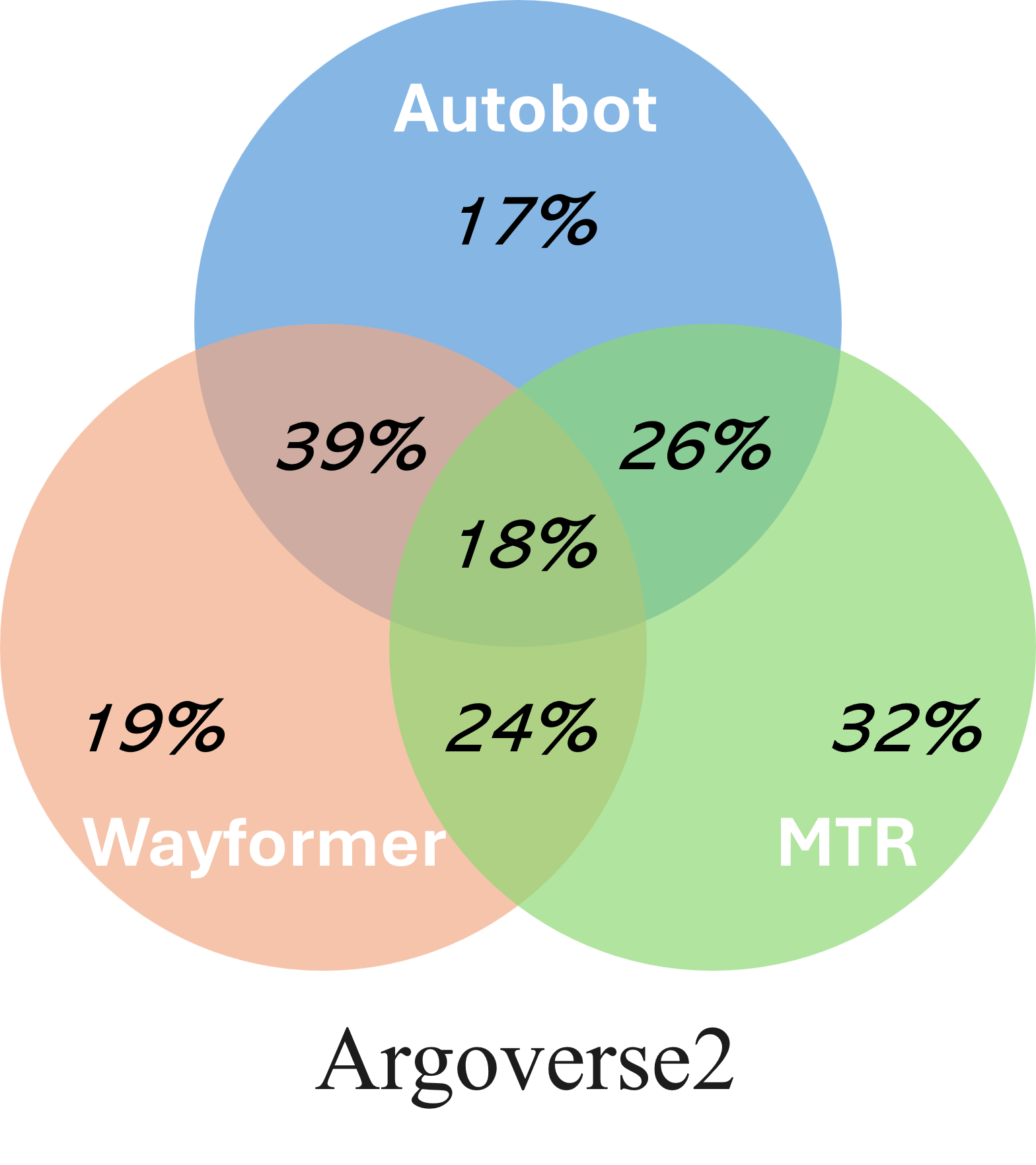}
    \caption{This figure shows the extent to which the set of long-tail examples (Top {10}\% most challenging examples, i.e. $\mathcal{D}_{\text{top-}10\%}$ for each model) overlap with each other on the NuScenes and Argoverse 2 datasets. 
    % This shows that the set of examples that each model has difficulties with is not the same as that of other models.
    }
    \label{fig:sample_distribution}
    \vspace{-15pt}
\end{figure}
Furthermore, in order to examine the long-tailed dataset differences between the three models, we analyze how much of the data which falls into the long tail (i.e. the $\mathcal{D}_{\text{top-}10\%}$) for each of the three models, is shared, as shown in Figure~\ref{fig:sample_distribution}. We find that the $\mathcal{D}_{\text{top-}10\%}$ for each model shares only 15-18\% of its examples with the set of long tail examples common to all three models, while around one third of the $\mathcal{D}_{\text{top-}10\%}$ for each model is not common to that of either of the other two models.
This shows that each model has different strengths and weaknesses, since, for example, AutoBot and Wayformer perform better on MTR's top 10\% most difficult examples, and vice versa for each of the three models -- on MTR's $\mathcal{D}_{\text{top-}10\%}$, AutoBot's ADE / FDE is 5.33 / 13.91 and Wayformer's is 4.90 / 13.77, which is much better than MTR's performance of 7.48 / 20.85.
% --  and vice versa for each of the three models.
This analysis helps explain why even the lower performing models help make the weighted average ensemble performance better.

% are common to the long tail of the other two models, whereas more than a third of 

% are different for every model, i.e.  This shows that the three models have different strengths and weaknesses and helps explain why even the lower performing models help make the weighted average ensemble performance better.

\subsection{Performance of Weighted Ensemble}

As shown in Tables~\ref{tab:nus} and~\ref{tab:av2}, our method, though simple, improves the metrics across the board when the outputs of multiple models are combined. When using an ensemble of two models, the performance shows consistent improvements across both datasets over the top mode performances of the individual models used. On the NuScenes dataset, the combined Autobot and Wayformer ensemble (16.5M parameters) even shows better performance than the pre-trained MTR model (60.1M parameters), which is almost 4 times bigger than the Autobot and Wayformer models combined, showing that on some datasets, ensembling can achieve better performance with fewer parameters. 
% no matter which combination of models is used, using 2 models consistently outperforms even the best of the 3 pre-trained models, and 

As expected, the biggest improvements on NuScenes are made by the combination of all three models together. Even though AutoBot performs much worse than the other models on NuScenes, its addition to the ensemble adds small improvements to the overall and long-tail performances, making the three-model ensemble even better than the two-model ensemble without Autobot. 
% Of note is the fact that the AB + WF ensemble in Table~\ref{tab:nus}, which has a combined 16.5M parameters, performs better than MTR, which has 60M parameters

On the Argoverse 2 dataset, however, the overall performance of all the ensemble models is quite similar, except for the AB + WF model which is significantly worse (see Table~\ref{tab:av2}). This result can be explained by the fact that MTR has much better performance than AutoBot and Wayformer due to the fact that it is 4 times bigger than Wayformer and 40 times bigger than AutoBot, and therefore has the model capacity to learn more from the much bigger Argoverse 2 dataset. Therefore, the smaller combined AB + WF ensemble cannot perform better than the bigger MTR model on this much bigger dataset. Furthermore, the rest of the ensemble models which include MTR perform only about 1-3\% better on FDE and 4\% better on ADE than MTR, which shows that they may be relying on MTR's confident predictions to form a disproportionately large portion of the weighted average.
As for the long-tail performance, long-tail metrics on the Argoverse 2 dataset show consistent improvement as we add more models to the ensemble: here we again see the 10\% improvement from the three-model ensemble that we saw in NuScenes. 
% These results come at a computational cost of 27.5\% more parameters than MTR, the best of the three models.

It is interesting to note that AutoBot, though it has 10x fewer parameters, is the second best-performing model on the Argoverse 2 dataset. This may be because we are using only the most likely prediction here. While Wayformer is good at predicting a wide distribution of good predictions, it's not as good at choosing the best of those predictions to be the most likely. 

For both datasets, long-tail metrics improved significantly when using the ensemble. As was discussed in \cite{filosCanAutonomousVehicles2020a}, when it comes to predicting out-of-distribution scenes, which long-tail examples commonly are, ensembles have been shown to help by reducing overconfident catastrophic extrapolation. In our case, we see that in long-tailed examples, variance between the three models' predictions is high, as can be seen in the top-right example in Figure~\ref{fig:nuScenes}, and by taking the weighted average of the predicted trajectories, we can find a result in the middle that can be closer to the ground truth. 

Most of the cases in which the ensemble shows the most improvement over the baseline are ones in which there is relative agreement among the models on whether the vehicle will go straight, right, or left, but high variance across the predicted paths taken, like in the top row of examples in Figure~\ref{fig:nuScenes}. Since our method simply performs a weighted average, in these cases, our method achieves better results than the individual models by finding consensus among them. 

\subsection{Failure cases}
% When the different models are predicting different modes (e.g. one model predicts the vehicle will go straight and the other predicts it will turn), this can lead to a weighted average that straddles the two modes, as can be seen in the bottom-right examples in Figures~\ref{fig:nuScenes} and~\ref{fig:av2}. This is one of the main failure modes of our weighted-average ensemble method, and can cause downstream tasks like path planners to expect trajectories that would never occur in real life. 

The failure cases of the proposed ensemble method typically fall into one of two scenarios: either one of the models is overconfident and wrong (as is the case in the bottom left examples in both Figures~\ref{fig:nuScenes} and~\ref{fig:av2}), or the three models are predicting different modes of behaviors (e.g. one model predicts the vehicle will go straight and the other predicts it will turn) and the ensemble experiences mode collapse by predicting the path in between. 

In the second case, this can lead to a weighted average that straddles the two modes, as can be seen in the bottom-right examples in Figures~\ref{fig:nuScenes}-~\ref{fig:av2}. This can cause downstream tasks like path planners to expect trajectories that would never occur in real life. 
However, our variance-based confidence score can help alert downstream tasks to the low quality of the prediction by translating the high variance between the three models' predictions into a low confidence score.

\subsection{Ablation}

\begin{table}[ht]
\centering
\setlength{\tabcolsep}{3pt}
\begin{tabular}{lcc|cc}
\toprule
\textbf{Method} & \multicolumn{2}{c|}{\textbf{nuScenes}} & \multicolumn{2}{c}{\textbf{Argoverse}} \\
 & Top 10\% & Overall & Top 10\% & Overall \\
\midrule
Simple             & 6.75 / 18.18 & 2.29 / 6.00 & 6.75 / 18.18 & 2.29 / 6.00 \\
{Weighted (Ours)}          & \textbf{6.68 / 18.10} & \textbf{2.25 / 5.91 }& \textbf{6.31 / 16.68} & \textbf{1.87 / 4.69} \\
Threshold-W~\cite{kimHybridApproachVehicle2021}  & 6.72 / 18.30 & 2.27 / 5.95 & 6.72 / 18.30 & 2.27 / 5.95 \\
\bottomrule
\end{tabular}
\caption{Ablation study: Comparison of ensemble strategies 
% (simple, weighted, and threshold-weighted averaging)
on nuScenes and Argoverse 2 datasets (ADE / FDE)}
\label{tab:ablation}
\vspace{-5pt}
\end{table}

To demonstrate the robustness of our results, we also compare our weighted average ensembling method with two other other types of averaging, as shown in Table~\ref{tab:ablation}. One is a simple unweighted average while the other is a thresholded weighted average proposed by~\cite{kimHybridApproachVehicle2021}, where thresholding is first applied on the confidence of the best performing model. If confidence is high, that model's output is used, otherwise the variance-weighted (or in our case, confidence-weighted) average over all models is used~\cite{kimHybridApproachVehicle2021}. Though this requires the extra step of determining the optimal threshold (which, for our ensemble, we found to be a confidence score of 0.75 on our best performing model, MTR, after trying thresholds at 0.25, 0.5, and 0.75), our weighted average method still outperforms both the simple and threshold-weighted average methods. This is likely because our confidence-weighted average automatically takes into account the relative confidences of all models instead of ignoring them or applying a preset threshold which does not adapt to each example. 

% such that the output of the best performing model in the ensemble is chosen without averaging if its confidence is above a certain threshold, otherwise confidence-weighted averaging over all the models is used.

\begin{table}[ht]
\centering
\begin{tabular}{lcc}
\toprule
\textbf{Method} & \textbf{Time (ms)} & \textbf{GFLOPs} \\
\midrule
AutoBot (AB)~\cite{girgis2022latent-autobot} & 20.98 & 1.55 \\
Wayformer (WF)~\cite{nayakanti2023wayformer} & 17.23 & 8.76 \\
MTR~\cite{shi2022motion} & 64.28 & 17.31 \\
Ensemble (Weighted) AB + WF + MTR & 100.63 & 27.62 \\
\bottomrule
\end{tabular}
\caption{Computational cost: inference time (ms) and computational complexity in giga-FLOPs (GFLOPs) of each method for one input example}
\label{tab:computational}
\vspace{-10pt}
\end{table}
% caption: Computational cost: time (ms) and compute (number of giga-FLOPs, GFLOPs) required to run each model on one example

\subsection{Computational Cost}
In addition to the memory requirements described in our methodology, we also report the computational cost of running each model compared to the combined ensemble. In Table~\ref{tab:computational} we show the number of floating-point operations (FLOPs) required to run each of the 3 base models, as well as our combined ensemble, on one input scene. This metric provides a hardware-independent way to estimate the computational complexity~\cite{flops}. 

% TODO - gpu
Furthermore, we also provide the time taken on our own GPU (NVIDIA A100 SXM4 40GB) to run one example through each of the base models, and through our ensemble (in which we run each base model serially, and then combine their results through the weighted average). Parallelization of the base models may yield a better computational time cost. 

\section{Conclusion and Future Work}

In this work, we provide a flexible approach to ensemble modeling which doesn't require re-training models, where newer state-of-the-art pre-trained models can be swapped in or out at any time. 
We use three previously state-of-the-art models perform a confidence-weighted average on their most likely predictions to produce an ensembled prediction which beats the best of the models' performances by around 10\%. 

In the future we aim to
expand to use more models. 
Instead of trying all possible combinations like in Tables~\ref{tab:nus}-~\ref{tab:av2}, we plan to automatically select for the optimal set of models using a method like \cite{salinasTabRepoLargeScale2024} which iteratively adds successive models such that the average of the selected models’ predictions minimizes overall error \cite{caruanaEnsembleSelectionLibraries2004}. 
Furthermore, since our method's biggest drawback is having to run multiple models at inference time, we also plan to explore knowledge distillations strategies like those in  \cite{gouKnowledgeDistillationSurvey2021} to train one model to achieve the same output as our ensemble. 

\section*{Acknowledgement}

This work was jointly supported by Vinnova, Sweden (research grant), the EPSRC Programme Grant ``From Sensing to Collaboration'' (EP/V000748/1), and RobotCycle. The computations were enabled by the supercomputing resource Berzelius provided by National Supercomputer Centre at Linköping University and the Knut and Alice Wallenberg Foundation, Sweden. 
%%%%%%%%%%%%%%%%%%%%%%%%%%%%%%%%%%%%%%%%%%%%%%%%%%%%%%%%%%%%%%%%%%%%%%%%%%%%%%%%

\bibliographystyle{IEEEtran}
\bibliography{IEEEabrv,ref}

\end{document}